\documentclass[letterpaper, 10 pt, conference]{ieeeconf}  

\IEEEoverridecommandlockouts                              

\usepackage{xcolor}
\usepackage{amsmath}
\usepackage{booktabs}
\usepackage{graphicx}
\usepackage{subcaption}
\usepackage{cite}
\usepackage{mathtools}
\usepackage{graphicx}
\usepackage{dsfont}
\usepackage{tikz}
\usepackage{mathrsfs}

\graphicspath{{img/}{../img/}}
\mathtoolsset{showonlyrefs}

\usepackage{subfiles} 

\title{\LARGE \bf
Safety Enhancement for Deep Reinforcement Learning in Autonomous Separation Assurance
}

\author{Wei Guo$^{1}$, Marc Brittain$^{2}$, Peng Wei$^{3}$
\thanks{$^{1}$Wei Guo is with the Department of Computer Science, George Washington University, Washington, DC 20052, USA, {\tt\small weiguo@gwu.edu}}%
\thanks{$^{2}$Marc Brittain is with the Department of Aerospace Engineering, Iowa State
University, Ames, IA, 50021, USA
        {\tt\small  mwb@iastate.edu}}%
\thanks{$^{3}$Peng Wei is with the Department of Mechanical and Aerospace Engineering, George Washington University, Washington, DC 20052, USA
        {\tt\small pwei@gwu.edu}}%
}

\begin{document}

\maketitle
\thispagestyle{empty}
\pagestyle{empty}

\begin{abstract}
The separation assurance task will be extremely challenging for air traffic controllers in a complex and high-density airspace environment. Deep reinforcement learning (DRL) was used to develop an autonomous separation assurance framework in our previous work where the learned model advised speed maneuvers. In order to improve the safety of this model in unseen environments with uncertainties, in this work we propose a safety module for DRL in autonomous separation assurance applications. 
The proposed module directly addresses both model uncertainty and state uncertainty to improve safety.
Our safety module consists of two sub-modules: (1)
the \textit{state safety sub-module} is based on the execution-time data augmentation method to introduce state disturbances in the model input state; (2) the \textit{model safety sub-module} is a Monte-Carlo dropout extension that learns the posterior distribution of the DRL model policy.
We demonstrate the effectiveness of the two sub-modules in an open-source air traffic simulator with challenging environment settings.
Through extensive numerical experiments, our results show that the proposed sub-safety modules help the DRL agent significantly improve its safety performance in an autonomous separation assurance task.
\end{abstract}

\section{Introduction}

\subsection{Motivation}


As a consequence of the growth of global air traffic, ensuring the safety and scalability of air traffic control (ATC) becomes a key challenge.
Since air traffic controllers' human cognition will not match up with the high traffic complexity and density, 
introducing automation tools to help human controllers guarantee the safe separation in these environments becomes a necessity. 
The Advanced Airspace Concept (AAC) paved the way in autonomous air traffic control by designing automation tools to assist human controllers to resolve pairwise conflicts \cite{erzberger2005automated,farley2007fast}. Motivated by the idea of the AAC, we trust that a highly automated ATC system is the solution to manage the high-density, complex, and dynamic air traffic in the future en route and terminal airspace.

Deep reinforcement learning (DRL) is a promising solution to build the  foundation for the envisioned autonomous ATC. 
The powerful DRL can be trained to learn compact low-dimensional representations of high-dimensional data (i.e., state space of aircraft) to reason about the environment and perform tactical decision making.
The authors have introduced and maintained the state-of-the-art in developing multi-agent DRL methods for autonomous separation assurance in structured airspace \cite{brittain2019autonomous, brittain2021one, brittain2020deep}. However, errors and accidents may happen in safety-critical systems when there exists  out-of-sample scenarios in execution time, and state and model uncertainties across training and execution \cite{farebrother2018generalization, cobbe2019quantifying,cobbe2020leveraging}. 
The objective of this work is to design a safety module to support the DRL agent under state and model uncertainties in execution phase.

\textcolor{black}{One approach to enhancing the  safety in DRL is to add state or action constraints to the reinforcement learning problem formulation. In constrained reinforcement learning, the agent maximizes the return while keeping some constraints satisfied \cite{kadota2006discounted}.} 
\textcolor{black}{However, in the multi-agent setting of this paper, the constrained RL problem formulation would depend on a large set of coupling constraints between agents, which makes the model training intractable.}

This work is built on top of our previous work \cite{brittain2020deep}. In this paper a complimentary safety module is designed and incorporated to the original DRL model during the execution phase. 
In contrast to the related papers, our proposed module provides safe and robust decision making under model uncertainty and state uncertainty. More importantly, it can be inserted into any pre-trained DRL network to provide immediate safety enhancement without further training or transfer learning.
Specifically, this proposed module, named the dropout and data augmentation (DODA) safety module, incorporates a \textcolor{black}{model safety sub-module to quantify the DRL model uncertainty based on MC-dropout and a state safety sub-module based on execution-time data augmentation (DA)} to handle the state uncertainty.

The main contributions of this paper can be summarized as follows:

\begin{itemize}
    \item We propose a safety enhancement module called DODA to improve the safety performance of DRL agents in safety-critical systems. This module can provide immediate safety enhancement to a general DRL agent without additional training or transfer learning. 
    \item The two sub-modules directly address state and model uncertainties. We show the ablation studies on the effectiveness of both sub-modules.
    \item We demonstrate the effectiveness of the integrated DODA safety module in   complex and challenging ATC environments with an open-source simulator, which outperforms the DRL agents without the DODA module. 
\end{itemize}


\subsection{Related Work}

\begin{figure*}[ht]
\centering
\label{fig:major}
\includegraphics[width=15cm, height= 5cm]{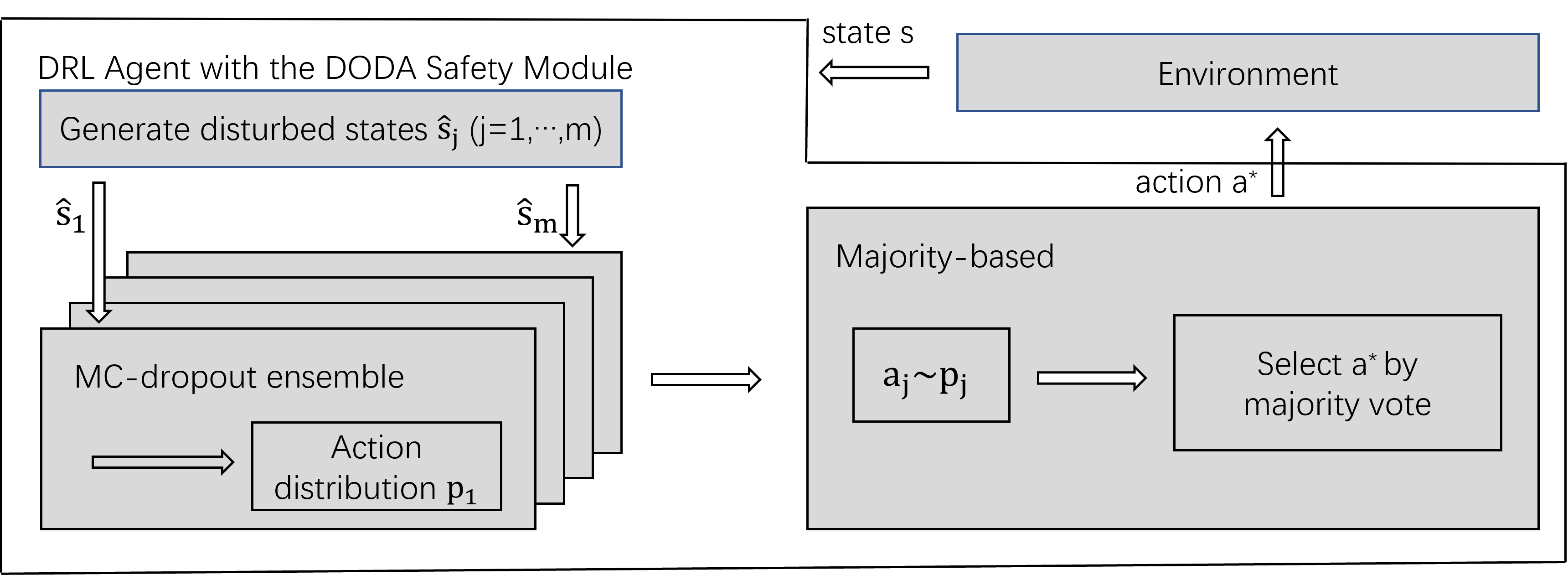}
\caption{Structure of the DODA safety module. An agent observes the environment and selects the safe action. At each time step, $m$ disturbed states are generated. An MC-dropout ensemble fits the action distribution $p$ for each disturbed state. Majority vote  criterion is used to select the final action $a^*$.}
\label{fig:flow}
\end{figure*}

DRL has been widely explored in ATC. 
During the recent years, the authors were one of the earliest to introduce DRL in autonomous ATC. The problem of enroute ATC automation in structured airspace using DRL was firstly addressed by Brittain and Wei
\cite{brittain2019autonomous} which was based on a deep neural network. 
We also proposed to use
Long-Short Time Memory (LSTM) network \cite{brittain2020deep}, and attention network \cite{brittain2021one} to handle a variable number of aircraft in separation assurance problem. Mollinga and Hoof \cite{mollinga2020autonomous} built a graph-based network for ATC in three-dimensional unstructured airspace.
Tran et al. \cite{tran2020interactive} employed the deep deterministic policy gradient to  perform vectoring. 
Dalmau and Allard \cite{dalmauair} introduced a message-passing actor-critic network to support separation assurance.
Ghosh et al. \cite{ghosh2020deep} built an combination of a local kernel-based and a deep multi-agent reinforcement learning model to dynamically suggest adjustments of aircraft speeds in real-time.
However, in these works, the authors focused on training a safe RL policy or a safe DRL agent, which is often vulnerable when out-of-sample scenarios are seen in execution time. This limits their usage in safety-critical systems such as autonomous ATC. In this work, we present a safety module which pairs with a DRL agent to significantly enhance its safety performance when facing out-of-sample execution scenarios with state and model uncertainties.

The idea of run-time monitoring has been firmly established and requires the specification of properties satisfied by the software or system, and the supervision of those properties during execution to ensure those are met over a partial trace. 
Furthermore, the monitoring system can offer a measure of the
``robustness'' with which the system satisfies (resp. violates) the specifications by evaluating how close the system is to violating (resp. satisfying) the specification \cite{deshmukh2017robust}. When well designed, such a monitoring component can be used to detect, or even predict unsafe or undesired behavior during the execution of a complex cyber-physical system. 

For safe guarding DRL models, a safe reinforcement learning framework \cite{lutjens2019safe} uses bootstrapping and MC-dropout to provide computationally tractable uncertainty estimates to bridge the gap between training scenarios and execution scenarios.
In addition, an online certified defense is designed in \cite{lutjens2020certified},
which computes guaranteed lower bounds on state-action values during execution to identify and choose the optimal action under a worst-case deviation in input space due to possible adversaries or noise.
\textcolor{black}{However, the construction of these bounds highly relies on the type of the adopted DRL algorithms and the bounds could be too conservative.}

This paper focuses on enhancing the safety of DRL models via addressing the 
model uncertainty and state uncertainty.
Model uncertainty here accounts for uncertainty in the deep reinforcement learning model parameters which can be reduced given enough training data or transitions \cite{kendall2017uncertainties}. 
State uncertainty captures noise inherent in the observations \cite{kendall2017uncertainties}. This could be the sensor noise or communication noise, which leads to the inaccurate input state to the DRL model.

To capture the model uncertainty, Bayesian neural networks (BNNs) are introduced \cite{denker1990transforming, mackay1992practical, neal2012bayesian}, which replace the deterministic network's weights with distributions over these weights. Since BNNs are computationally expensive to perform inference, dropout variational inference is introduced as a practical approach for approximate inference \cite{gal2015bayesian}, which is done by performing execution-time dropout to sample from the approximate posterior distribution and referred as MC-dropout. 
MC-dropout is widely used in DRL to quantify the model uncertainty \cite{gal2016improving, kahn2017uncertainty, malik2019calibrated,lutjens2019safe}. 
Another line of research uses a Hypernet \cite{pawlowski2017implicit} that uses another network to give parameter uncertainty values. It has been shown to be computationally intractable.
In this work, MC-dropout is used to estimate the model uncertainty.

State uncertainty is often introduced by noise from observation, sensing, or communication. The data augmentation (DA) method was first proposed in computer vision to expand a small dataset by applying transformations to its samples during the training phase \cite{krizhevsky2012imagenet}. 
Recent work has shown that the use of DA is beneficial in the training process of DRL \cite{cobbe2019quantifying, laskin2020reinforcement} to improve the generalization. 
Ayhan and Berens \cite{ayhan2018test}  found that execution-time DA
can be used to estimate the state uncertainty in deep networks. 
Our work extends the execution-time DA to DRL application in autonomous separation assurance. 
 

The remaining parts of the paper are organized as follows. Section~\ref{sec:2} introduces the background. In Section~\ref{sec:3}, the description of the problem is  introduced. Section~\ref{sec:4}
introduces our  solution to the problem. Section~\ref{sec:5} and Section~\ref{sec:6} give the details of the experiments and results.
Section~\ref{sec:7} summarizes our findings and concludes the paper.

\section{Background}
\label{sec:2}

\subsection{Monte Carlo Dropout}

MC-dropout is a practical approach for approximate inference in large models \cite{gal2015bayesian}. 
The process is derived by training a model with dropout and performing execution-time dropout to sample from the approximate posterior distribution. 

Theoretically, this method is equivalent to performing approximate variational inference where we try to find a tractable distribution $q_{\theta}^{*}(W)$ which minimizes its KL divergence to the true model posterior distribution $p(W|X,Y)$. 
Then the minimization objective \cite{jordan1999introduction} is given as follows with $N$ samples, dropout probability $p$, samples $\hat{W}_i \sim q_{\theta}^* (W)$, and the parameter set $\theta$ of the tractable distribution:
\begin{equation}
    L(\theta, p) = -\frac{1}{N} \sum_{i=1}^N \log p(y_i|f^{\hat{W_i}}(x_i)) + \frac{1-p}{2N} \vert\vert \theta \vert\vert^2.
\end{equation}
Model uncertainty can be reduced by observing more data. For classification, this can be approximated using Monte Carlo integration as follows with $n$ sampled masked model weights $\hat{W}_t \sim q_{\theta}^* (W)$, and the dropout distribution $q_{\theta}(W)$:
\begin{equation}
    p(y=c|x,X,Y) \approx \frac{1}{n} \sum_{i=1}^n \text{softmax}(f^{\hat{W}_t}(x)).
\end{equation}

\subsection{Execution-Time Data Augmentation}
Execution-time DA was proposed in computer vision for model generalization. 
The augmentation methods
typically include flipping, cropping, rotating, and scaling training
images.
Several
studies have empirically found that combining outputs of
multiple transformed versions of a test image helps to improve
the model performance \cite{matsunaga2017image, radosavovic2018data}. 

In our case, we implement the data augmentation with the state information (such as position, velocity, etc.) of the aircraft instead of images. We add a noise $\epsilon \sim \text{U}(-0.1,0.1)$ to the normalized input state value and then clip it to the range between 0 and 1.
Uniform distribution is used since we assume that the noise values  all have equal probability.
Our approach amounts to generating $m$ augmented examples per case and feeding them to the DRL model to obtain a distribution of actions. This method selects the output action following this action distribution.

\section{Problem Formulation}
\label{sec:3}

In en route sectors, air traffic controllers are responsible for the safety separation among all aircraft. In this work, we use the BlueSky \cite{hoekstra2016bluesky} as the air traffic simulator. 
We evaluate the performance of our \textcolor{black}{proposed execution-time safety module DODA} on several challenging case studies with out-of-sample scenarios, unseen environments, multiple intersections, and high-density air traffic.


The objective of the case studies is to maintain safe separation between aircraft and resolve conflicts for all aircraft
\textcolor{black}{in an unseen sector} by providing speed advisories.
The agents need to plan and coordinate with other agents before the intersection to ensure the safety separation since commercial aircraft are unable to hover or stop in front of the intersection.
Four different \textcolor{black}{cases} are used in this work: case A is used as the training environment for the DRL agent to obtain a converged policy, and the other three cases are used to evaluate the policy's performance and the proposed DODA safety module's effectiveness in unseen environments with state/model uncertainties. 
In these four case studies, the training case A and the three evaluation cases B, C, D are very different, which is challenging for the agents in the simulation environment since the policy converged in case A is not safe anymore for the other three cases.




\textbf{Multi-Agent Reinforcement Learning Formulation}. We formulate the safety assurance problem as a deep multi-agent reinforcement learning problem by treating each aircraft as an agent. We define the state space, action space, termination criteria, and reward function as follows.

\subsubsection{State Space}
The state contains all the information an agent needs for decision making.
Since our work is a cooperative multi-agent setting where aircraft do not want to have collisions with others, we implement the communication and coordination between the agents. 
Specifically, the state information of the ownship includes the following values: current location, the distance to the goal (the exit of this sector), current speed, acceleration, heading of the aircraft. 
And the state information of each intruder includes the following items:
current location, current speed, acceleration, heading of the intruder, distance from ownship to the intruder, distance from ownship to intersection, distance from the intruder to intersection.
The information of all intruders are collected and used as the input state to the attention network (see \cite{brittain2020deep} for details), which allows the ownship to have access to the information of all intruders. 


\subsubsection{Action Space}
We allow each agent to select the action every 12 seconds (en route position update is available every 12 seconds from radar surveillance). The action space of each agent is defined as follows:
\begin{equation}
    \mathcal{A}_t = [a_{-},a_{0},a_{+}].
\end{equation}
Here $a_{-}$ refers to deceleration, $a_{0}$ refers to maintaining the current speed, and $a_{+}$ stands for acceleration.

\subsubsection{Terminal State}
In each simulation, aircraft will be generated in the sector until the maximum number of aircraft is reached. The simulation will then terminate when all aircraft have reached their individual terminal state, obtained in the following two ways: (1) violating loss of separation, (2) exiting the sector without conflict.

\subsubsection{Reward Function}
We define identical reward function for all agents to encourage the cooperation. 
The system penalizes the conflict with a negative reward: the penalty is local that only the two or multiple aircraft in conflict will receive the penalty, but the other agents will stay unaffected.
In this work, a conflict is defined as the distance between two aircraft is less than 3 nautical miles. Besides the penalty of conflicts, we introduce the penalty of speed changes, which in a real world setting should be avoided unless necessary. We capture our goals in the reward function as follows:
\begin{equation}
    R(s,a) = R(s) + R(a).
\end{equation}

$R(s)$ and $R(a)$ are defined as:
\begin{equation}
R(s)=
\begin{cases}
-1 & \text{if} \: d_o^{c} < 3\\
-\alpha + \delta \cdot d_o^{c} &  \text{if} \: 3 \leq d_o^{c} < 10 \\
0 & \text{otherwise}
\end{cases}
\end{equation}

\begin{equation}
    R(a) = 
    \begin{cases}
    0& \text{if} \: a = a_0\\
    -\psi & \text{otherwise.}
    \end{cases}
\end{equation}

Here $d_o^{c}$ is the distance from the ownship to the closet intruder in nautical miles. $\alpha$ and $\delta$ are used to ensure the reward is between -1 and 0.
The reward function $R(s)$ allows the agent to learn a policy whose goal is to ensure safe separation. $\psi$ is implemented to minimize the number of speed changes.

\section{Solution Approach}
\label{sec:4}

Our objective is to improve the safety performance of the DRL agent in autonomous separation assurance. In order to achieve this goal, we design an execution-time safety enhancement module to pair with the DRL model. The system consists of two safety sub-modules, namely, the \textit{model safety sub-module} and \textit{state safety sub-module}.
The model safety sub-module is an ensemble based on MC-dropout. 
The state safety sub-module creates disturbed input states into the system and finds the most robust decision given the noisy input states. We describe two sub-modules in detail in the following subsections and then show how they are integrated as the DODA execution-time safety module.

\subsection{Model Safety Sub-module}




The model safety sub-module is based on MC-dropout \cite{gal2016dropout}.
MC-dropout method randomly deactivates network units in each forward pass by multiplying the unit weights with a dropout mask.
The dropout mask is a set of Bernoulli random variables of value 0 and 1, each with the same probability.
Our sub-module executes multiple forward passes per input state with different dropout masks and acquires a distribution over actions. 

For each input state $s$, a total of $n$ forward passes are sampled. Given our trained DRL policy network with dropout $\pi$, we obtain $n$ actions for the state $s$ as:
\[
a^{1}(s),...,a^n(s).
\]
Based on the sampled actions, an action distribution $p$ is generated as follows:
\[
\text{P}(a) = \frac{1}{n} \sum_{i=1}^{n} \mathds{1}_{a}(a^{i}(s))
\]
where $\mathds{1}$ is an action indicator.
Then the final action $a^*$ will be selected following the distribution $p$.

\subsection{State Safety Sub-module}
The model safety sub-module is not sufficient to prevent collisions in unseen environments because the state uncertainty has not been directly addressed yet. 
Therefore, we implement a state safety sub-module based on execution-time data augmentation (DA). To handle the state uncertainty introduced by sensors, communication and other sources, this sub-module intentionally adds simulated random noise $\epsilon \sim \text{U}(-0.1,0.1)$ to the normalized state $s$. Then the sub-module limits the value of noisy state $\hat{s}$ between 0 and 1.
Thus the disturbed state is constructed as follows:
\begin{equation}
    \hat{s}_j = \text{clip}(s + \epsilon), j=1,...,m, \epsilon \sim \text{U}(-0.1,0.1),
\end{equation}
where $j$ denotes that it is $j$th disturbed state vector for the state vector $s$. 
Each disturbed state is used as the input state to the DRL model for action selection.


Given the disturbed states, majority-based method is applied in state safety sub-module.
For each disturbed state $\hat{s}_j\:(j=1,...,m)$, one forward pass is sampled with the DRL policy $\pi$. We obtain $m$ actions as:
\[
a(\hat{s}_1),...,a(\hat{s}_m).
\]
State safety sub-module equally considers the selected actions from $m$ disturbed states and performs the majority vote to select the final action $a^*$ as follows:
\[
a^* = \operatorname*{argmax}_{a\in \mathcal{A}_t} \sum_{j=1}^m \mathds{1}_a (a(\hat{s}_j)).
\]

\subsection{Integration of Two Sub-modules}

To address both the model uncertainty and state uncertainty simultaneously in the execution phase, we integrate the model safety sub-module and the state safety sub-module together. We illustrate the architecture of the integrated module in Figure~\ref{fig:flow}.

The integrated module is built \textcolor{black}{mainly} based on the structure of state safety sub-module. The state safety sub-module generates $m$ disturbed states  $\hat{s}_j \: (j=1,...,m)$, and the model safety sub-module is used to generate the action distribution $p_j$ for each disturbed state.

For each disturbed state $\hat{s}_j$, the model safety sub-module samples $n$ forward passes with MC-dropout and generates an action distribution $p_j$. Then $m$ disturbed states $\hat{s}_j$ and the corresponding action distributions $p_j$ are used in the  state safety sub-module.

In state safety sub-module, one action is generated following the action distribution for each disturbed state.
The majority vote is implemented to select the final action $a^*$.

\subsection{Baseline}
To demonstrate the effectiveness of our DODA safety module, we compare a DRL agent without the DODA and the same DRL agent with the DODA. Specifically, we use D2MAV-A \cite{brittain2020deep} as our DRL agent baseline. The D2MAV-A DRL model currently holds the state-of-the-art in autonomous separation assurance in structured airspace.
In this work, a dropout mask is added to two hidden fully-connected layers in the D2MAV-A to enable the dropout.

\section{Numerical Experiments}
\label{sec:5}

\subsection{Simulator}

\begin{figure*}[ht!]
\centering
\begin{subfigure}{.45\linewidth}
    \centering
 \includegraphics[width=\textwidth, height=3.5cm]{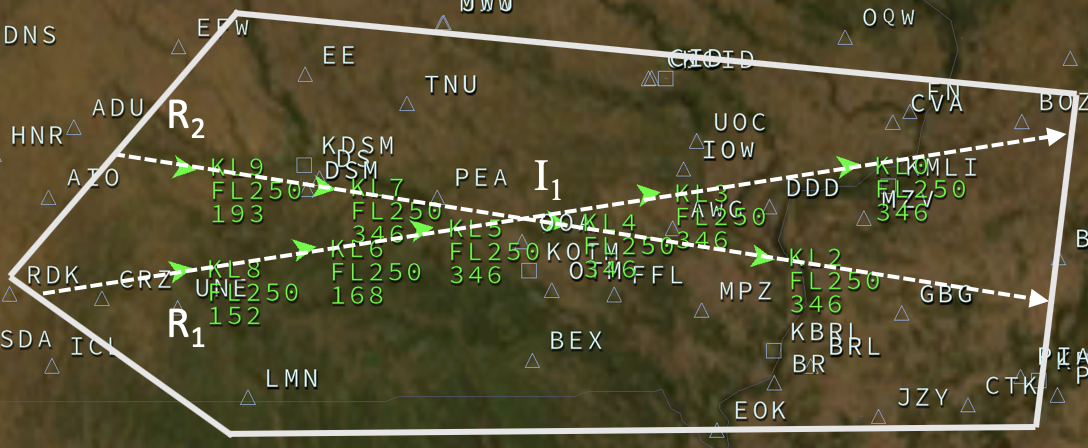}
         \caption{Case A is used only for DRL model training}
\end{subfigure}
    \hfill
\begin{subfigure}{.45\linewidth}
    \centering
   \includegraphics[width=\textwidth, height=3.5cm]{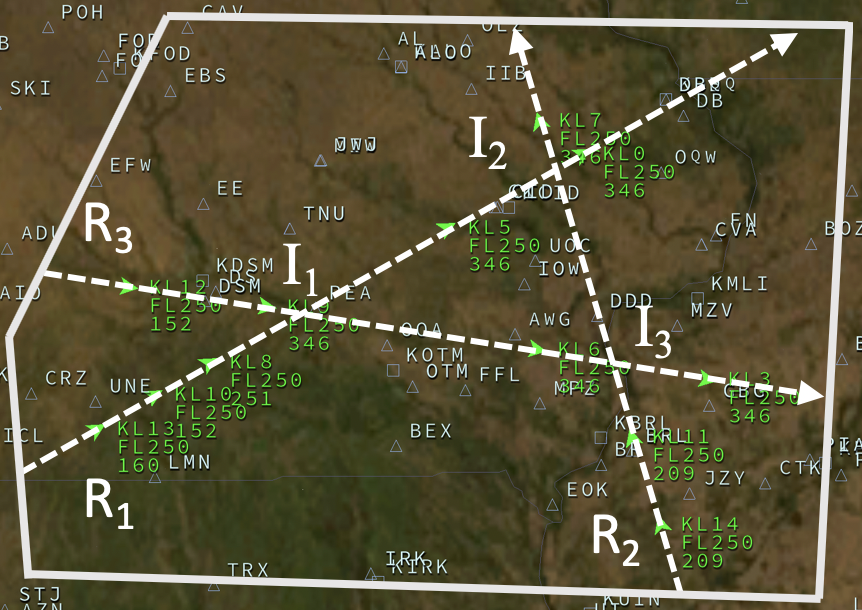}
         \caption{Case B is used for DRL and DODA validation}
\end{subfigure}

\bigskip
\begin{subfigure}{.45\linewidth}
  \centering
         \includegraphics[width=\textwidth, height=3.5cm]{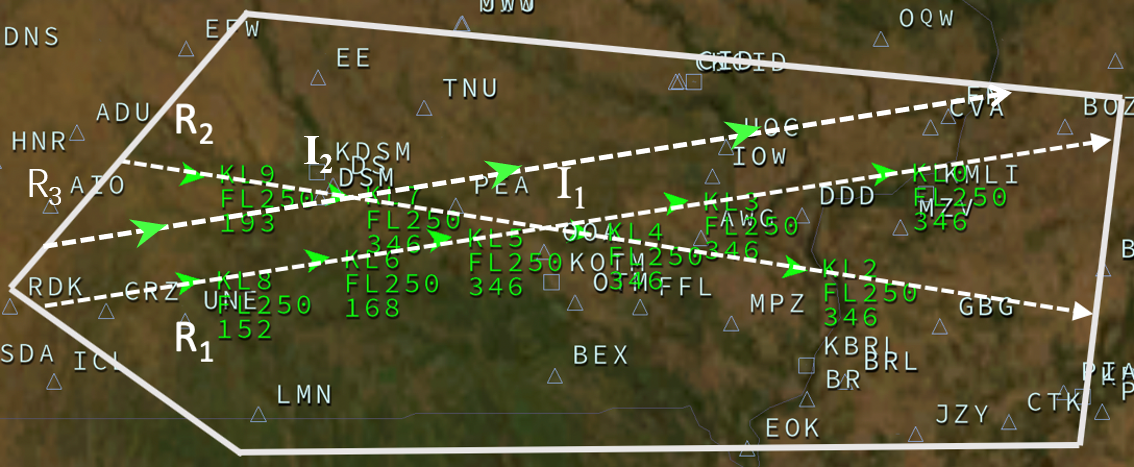}
         \caption{Case C is used for DRL and DODA validation}
\end{subfigure} 
    \hfill
\begin{subfigure}{.45\linewidth}
    \centering
         \includegraphics[width=\textwidth, height=3.5cm]{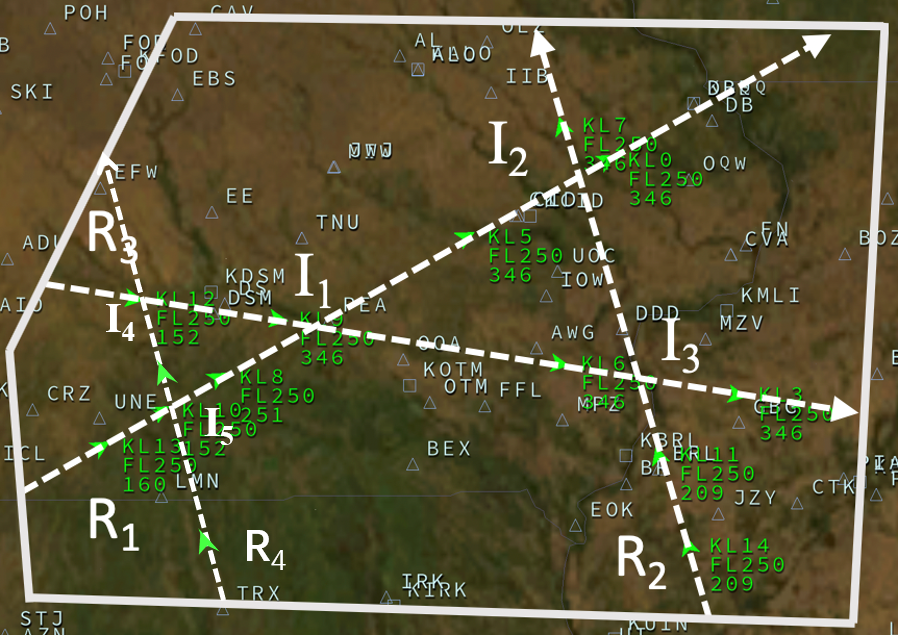}
         \caption{Case D is used for DRL and DODA validation}
\end{subfigure}
\caption{\textcolor{black}{Case Studies to train (in A) and validate (in B, C, D) the DRL model. Cases A and B are from \cite{brittain2020deep}.}}
\label{fig:case}
\end{figure*}

We utilize the BlueSky air traffic simulator built in Python \cite{hoekstra2016bluesky} to evaluate the safety performance of our proposed DODA module.
We can easily obtain the state information of all aircraft with the BlueSky simulator.

\subsection{Environment Settings}

We refer to each simulation run in BlueSky environment as a simulation: each simulation begins with only one aircraft on each route and ends when all aircraft have reached their terminal states.

Our safety module mainly consists of two sub-modules: the state safety sub-module, and the module safety sub-module. We introduce the important parameters in the safety module that we tune and select. We list the key parameter values in Table~\ref{table: safe-param}.

\begin{table}[ht]
\caption{Parameters of the DODA safety module.}
\label{table: safe-param}
\centering
\begin{tabular}{@{}ll@{}}
\toprule
Parameter & Value \\ \midrule
Drop probability & 0.2 \\
Size of MC-dropout & 5 \\
Size of DA & 5 \\
Distribution of noise & U(-0.1,0.1) \\
Training case & Case A \\ \bottomrule
\end{tabular}
\end{table}


For the parameters in the neural network, we follow the setting in D2MAV-A paper \cite{brittain2020deep}.


\subsection{Case Studies}

In this work, we consider four case studies \textit{A, B, C, D} with different routes and intersections \textcolor{black}{as shown in Figure~\ref{fig:case}}. 
The termination of simulations is when all aircraft in the airspace have reached their terminal states. Our goal is that every aircraft exits the sector without any conflict. 
These are very challenging problems to solve since the agents cannot memorize actions and must learn a strategy to achieve optimal performance (here we define ``optimal" as all aircraft that enter the sector must exist without any conflict).
 We define the reported score as the number of aircraft
exiting the sector without conflict. 

In our settings, the same neural network is implemented on each aircraft. Each aircraft selects its own desired speed based on the trained policy. This decentralized execution setting increases the complexity of this problem since the agents need to learn how to cooperate in order to maintain safe separation.

\subsection{Evaluation of Safety Performance}


In this experiment, we evaluate the effectiveness of our safety module in unseen case studies which the DRL model has not been trained on. 
Specifically, we train the DRL model on simplest case study A and evaluate the performance of the converged DRL model in other three more complex cases without any further training or policy updating. 
Case A is simplest since it has fewer routes or interactions than other three cases.
This is a difficult problem because the airspace structures of the other three cases are quite different. 
During the execution phase in cases B, C, and D which are unseen by the DRL agent, we evaluate the safety enhancement brought by our proposed DODA safety module.

Since our safety module has two sub-modules, we further implement a series of \textit{ablation experiments} which only use one sub-module to analyze their effectiveness: (1) only model safety sub-module used in execution phase; (2) only state safety sub-module used in execution phase; 
(3) both model safety and state safety sub-modules used in execution phase; (4) baseline DRL model without any execution-time safety module or sub-module.

In order to investigate the performance of the proposed safety module with different levels of air traffic density, we design two additional experiments with \textit{various air traffic densities} for each case study: one where the inter-arrival time interval is 2.6 to 3 minutes (high-density traffic), the other one where the inter-arrival time interval is 6 to 10 minutes (low-density traffic). It is important to notice that the difficulty of each case study is also based on the inter-arrival time of the aircraft, since the inter-arrival time controls the air traffic density in the airspace. The default interval in our ablation experiments is 3 to 6 minutes.

The output policy from the baseline DRL model, which is trained on case A without DODA module, is used to provide a fair comparison in our experiment.

\section{Result Analysis}
\label{sec:6}

In this section, we investigate the performance of the proposed safety module on 3 case studies: case B, case C, case D.
To compensate the stochasticity and provide a fair comparison, we report the average performance in 50 simulations for all experiments.

\subsection{Unseen Environments}

\begin{table}[t!]
\caption{Performance of the DODA module in unseen environments.}
\centering
\begin{tabular}{@{}lllll@{}}
\toprule
 & $\text{DA}$ &  DO & $\text{DA}$+DO  & Baseline \\ \midrule
Case B & \textbf{28.60}  & 27.21 & 27.60 &  24.60 \\
Case C & 29.93 & \textbf{30.00} & 29.81 &  25.69 \\
Case D & 25.40  & \textbf{26.20}  & 25.05 & 23.48 \\ \bottomrule
\end{tabular}
\label{table:ablation}
\end{table}

We report the performance of the DRL model extended by separate sub-modules or the integrated DODA safety module on three case studies in Table~\ref{table:ablation}.
Based on the last three columns in Table~\ref{table:ablation}, the DRL model enhanced by the DODA module obtains a higher score on all cases throughout the 50-simulation execution phase. 
The improvement achieved by the DODA module differs in each case study. 
We notice that our safety module works better in case B and case C than in case D.

Based on the first two columns in Table~\ref{table:ablation}, we find that each separate sub-module can still work to enhance the safety performance of the DRL model.
We do not find a clear pattern that one sub-module dominates the other one.

It is not certain whether the DODA safety module can still provide performance enhancement to DRL applications other than autonomous separation assurance. We will investigate this in our future work using more common validation platforms such as OpenAI Gym.
We will also explore the case  where the z-axis is taken into consideration in the future.


\subsection{Various Traffic Density}

\begin{table}[ht!]
\centering
\caption{Performance of the DODA module in various traffic densities.}
\begin{tabular}{@{}lllll@{}}
\toprule
  & \multicolumn{2}{l}{High Density} & \multicolumn{2}{l}{Low Density} \\ \cmidrule(l){2-5} 
  & DA+DO           & Base           & DA+DO          & Base           \\ \midrule
B & 26.40           & 22.60          & 27.83          & 27.40          \\
C & 29.20           & 18.80          & 30.00          & 28.00          \\
D & 24.40           & 20.87          & 26.80          & 26.17          \\ \bottomrule
\end{tabular}
\label{table:interval}
\end{table}

We report the performance of our DODA safety module with different traffic densities of the airspace on three case studies (B, C, D) in Table~\ref{table:interval}.
Based on the first  column in Table~\ref{table:interval},
we find that our safety module is able to enhance the safety of the DRL model in the execution case with \textcolor{black}{ a high density (2.6 to 3 minutes)} while the DRL model is trained in case A with a default density (3 to 6 minutes).
Based on the third column, we notice that the safety module also improves the performance in the different case study with a low density (6 to 10 minutes) if it is trained in case A with a default density (3 to 6 minutes). 
Comparing the performance on the same case with different densities, we notice that the safety module shows more significant improvement in the high-density case. The main reason is that the baseline DRL performs poorly when it is trained with a default density and then executed with a high density.


Furthermore, similar as the performance with the default density, the performance of the DODA module varies on three case studies. The best performance is observed in case C. The reason may be related to the different  airspace structure of each case study.


\section{Conclusion}
\label{sec:7}

In this paper, we propose a safety module for deep reinforcement learning which is used for autonomous air traffic control in a structured en route airspace sector. We observe that Monte-Carlo dropout and execution-time data augmentation can directly offset the uncertainties in unseen environments, thus can bring the immediate safety enhancement for DRL models without further training or policy updates. \textcolor{black}{This safety module will be useful in real-world safety-critical systems where transfer learning is needed from the DRL agent to adapt to new unseen environments, where our module will provide considerable amount of immediate safety enhancement before the DRL agent fully adapts.}
The promising results encourage us to further explore the effectiveness of the DODA module and its extensions for other safety-critical applications in the future.


\bibliographystyle{IEEEtran}
\bibliography{IEEEabrv,ref}

\end{document}